\def\year{2022}\relax
\documentclass[letterpaper]{article} % DO NOT CHANGE THIS
\usepackage{aaai22}  % DO NOT CHANGE THIS
\usepackage{times}  % DO NOT CHANGE THIS
\usepackage{helvet}  % DO NOT CHANGE THIS
\usepackage{courier}  % DO NOT CHANGE THIS
\usepackage[hyphens]{url}  % DO NOT CHANGE THIS
\usepackage{graphicx} % DO NOT CHANGE THIS
\usepackage{natbib}  % DO NOT CHANGE THIS AND DO NOT ADD ANY OPTIONS TO IT
\usepackage{caption} % DO NOT CHANGE THIS AND DO NOT ADD ANY OPTIONS TO IT

\DeclareCaptionStyle{ruled}{labelfont=normalfont,labelsep=colon,strut=off} % DO NOT CHANGE THIS
\newcommand{\etal}{\textit{et al}.}
\newcommand{\ie}{\textit{i}.\textit{e}.}
\newcommand{\eg}{\textit{e}.\textit{g}.} 

\usepackage{xcolor}
\usepackage{amssymb}% http://ctan.org/pkg/amssymb
\usepackage{pifont}% http://ctan.org/pkg/pifont
\newcommand{\cmark}{\ding{51}}%
\newcommand{\xmark}{\ding{53}}%

\usepackage{newfloat}
\usepackage{listings}
\usepackage{url}
\usepackage{booktabs}
\usepackage{nicefrac}
\usepackage{microtype}
\usepackage{epsfig}
\usepackage{subcaption}
\usepackage{amsmath}
\usepackage{amssymb}
\usepackage{amsthm}
\usepackage{bm}
\usepackage{multirow}
\usepackage{makecell}
\usepackage{footmisc}
\usepackage{marvosym}
\usepackage{latexsym}
\usepackage[utf8]{inputenc}
\usepackage{kotex}
\usepackage{bbm}
\usepackage{algorithm,algpseudocode}

\title{PrivateSNN: Privacy-Preserving  Spiking Neural Networks }
\author{
    Youngeun Kim, Yeshwanth Venkatesha, Priyadarshini Panda 
}
\affiliations{
Department of Electrical Engineering\\
Yale University\\
New Haven, CT, USA\\
{\{youngeun.kim, yeshwanth.venkatesha, priya.panda\}@yale.edu}
}

\usepackage{bibentry}
\begin{document}

\maketitle

\begin{abstract}
 How can we bring both privacy and energy-efficiency to a neural system? 
    In this paper, we propose PrivateSNN, which aims to build low-power Spiking Neural Networks (SNNs) from a pre-trained ANN model without leaking sensitive information contained in a dataset.
    Here, we tackle two types of leakage problems:
   1) Data leakage is caused when the networks access real training data during an ANN-SNN conversion process.
   2) Class leakage is caused when class-related features can be reconstructed from network parameters.
   In order to address the data leakage issue, we generate synthetic images from the pre-trained ANNs and convert ANNs to SNNs using the generated images.
   However, converted SNNs remain vulnerable to class leakage since the weight parameters have the same (or scaled) value with respect to ANN parameters.
   Therefore, we encrypt SNN weights by training SNNs with a temporal spike-based learning rule.
   Updating weight parameters with temporal data makes SNNs difficult to be interpreted in the spatial domain. We observe that the encrypted PrivateSNN eliminates data and class leakage issues with a slight performance drop (less than $\sim$2\%) and significant energy-efficiency gain (about 55$\times$) compared to the standard ANN.
    We conduct extensive experiments on various datasets including  CIFAR10, CIFAR100, and TinyImageNet, highlighting the importance of privacy-preserving SNN training. 
\end{abstract}

\section{Introduction}

Neuromorphic computing has gained considerable attention as an energy-efficient alternative to conventional Artificial Neural Networks (ANNs) \cite{he2016deep,simonyan2014very, christensen20222022, roy2019towards}.
Spiking Neural Networks (SNNs) process binary spikes through time like the human brain, and have been shown to yield 1-2 orders of magnitude energy efficiency over ANNs on emerging neuromorphic hardware \cite{roy2019towards,furber2014spinnaker,akopyan2015truenorth,davies2018loihi}.
Due to the energy advantages and neuroscientific interest, SNNs have made great strides on various applications such as image recognition \cite{lee2016training,kim2020revisiting,diehl2015unsupervised}, optimization \cite{fang2019swarm,frady2020neuromorphic},  object detection \cite{kim2019spiking}, and visualization \cite{kim2021visual}.
Going forward, SNNs offer a huge potential for power constrained edge applications \cite{venkatesha2021federated}.

\begin{figure}[t]
\begin{center}
\def\arraystretch{0.5}
\begin{tabular}{@{}c@{\hskip 0.08\linewidth}c@{}c}
\includegraphics[width=0.35\linewidth]{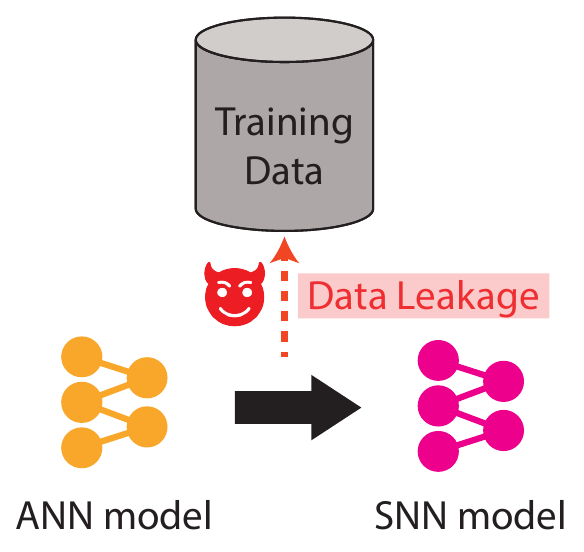} &
\includegraphics[width=0.35\linewidth]{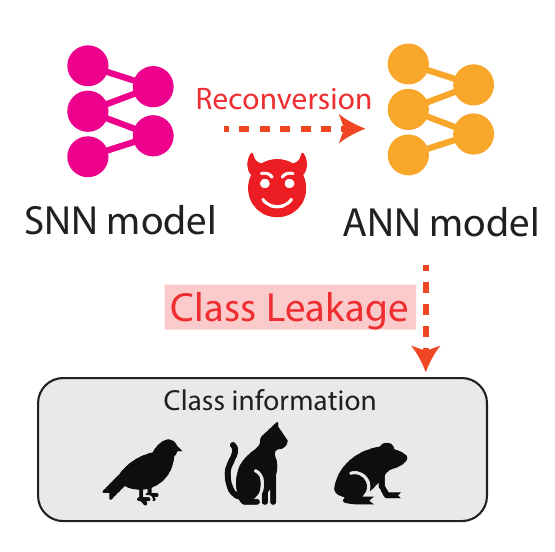} 
\\
% \vspace*{0.1in}
%  \vspace{-3.5mm}
{\hspace{-1mm} (a) During Conversion } & {\hspace{-1mm} (b)  After Deployment }\\
\end{tabular}
\end{center}
\caption{Illustration of data leakage and class leakage problems.
(a) The data leakage problem is likely to happen when an ANN model accesses real data during conversion process.
(b) The malicious attacker can obtain class information by reconverting  the SNN model to the ANN model.
}
 \vspace{-3mm}
\label{fig:intro:twodataleak_concept}
\end{figure}

Among various SNN training algorithms, ANN-SNN conversion is well-established and achieves high performance on complex datasets.
Here, pre-trained ANNs are converted to SNNs using weight or threshold balancing in order to replace Rectified Linear Unit (ReLU) activation with a spiking Leak-Integrate-and-Fire (LIF) activation \cite{sengupta2019going,han2020rmp,diehl2015fast,rueckauer2017conversion,deng2022temporal,li2021differentiable,kim2021optimizing}.
Existing conversion algorithms are based on the assumption that the model can access the entire training data. Specifically,  training samples are passed through a network and maximum activation value is used to calculate the layer-wise threshold or weight scaling constant.
However, this may not always be feasible.
Enterprises would not allow proprietary information to be shared publicly with other companies and individuals.
Importantly, the training set may contain sensitive information, such as biometrics.
Overall, such concerns motivates a line of research on privacy-preserving algorithms \cite{kundu2020universal,nayak2019zero,haroush2020knowledge,liang2020we}.
We refer to this problem as \textbf{data leakage} (Fig. \ref{fig:intro:twodataleak_concept}(a)).

\begin{figure}[t]
     \centering
         \includegraphics[width=0.35\textwidth]{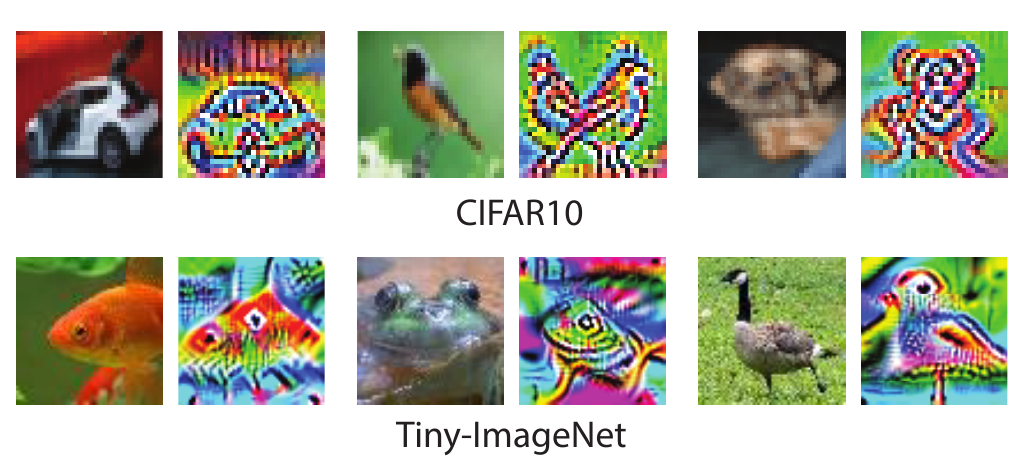}
% \vspace{1mm}
%  \vspace{-4mm}
 \caption{
Examples of class leakage.
We use VGG16 trained on CIFAR10 and Tiny-ImageNet. We visualize pair of images (left: real image, right: generated image) for each sample (CIFAR10: automobile, bird, and dog; Tiny-ImageNet: goldfish, bullfrog, and goose). The right image is generated using Algorithm 1.
 }
 \vspace{-5mm}
     \label{fig:intro:classleak}
\end{figure}

In addition, we tackle the \textbf{class leakage} problem  after deployment (\ie, inference), as shown in Fig. \ref{fig:intro:twodataleak_concept}(b).
It is a well-known fact that one can obtain a representative class image from model parameters by using simple gradient backpropagation (see Fig. \ref{fig:intro:classleak}) \cite{mopuri2018ask,yosinski2015understanding}.
From a security perspective, revealing class information can induce critical threats in a neural system.
A malicious attacker can find a blind spot in neural systems, and use these unrecognizable classes/pattern to disguise a system in real-world.
Also, class information can be exploited to generate a strong adversarial attack \cite{goodfellow2014explaining}. For instance, Mopuri \etal \cite{mopuri2018ask} use synthetic class representation to generate universal adversarial perturbations.
Therefore, to build a secure neural system, the class leakage problem should be addressed.

In this paper, we propose \textit{PrivateSNN}, a new ANN-SNN conversion paradigm that addresses both data leakage and class leakage problems.
Firstly, to address the data leakage issue, we generate synthetic data samples from the pre-trained ANN model and conduct ANN-SNN conversion using generated data.
For the class leakage issue, we encrypt the weight parameters with a temporal spike-based learning rule.
This stage optimizes the SNN parameters  with the non-differentiable spiking LIF activation function, which prevents exact backward gradients calculation. 
Note, it is difficult to address the class leakage problem in ANN domain since precise backward gradients can be calculated. 
At the same time, considering the resource-constrained devices where SNNs are likely to be applied, SNNs might be limited to using a small number of training samples due to the computational cost for training.
To preserve the performance with a small dataset, we distill the  knowledge from ANN to regularize the SNN during spike-based training.

In summary, our key contributions are as follows: 
(i) So far, in SNN literature, there is no discussion about the privacy issue of  ANN-SNN conversion.
For the first time, we showcase the privacy issues and also propose ways to tackle them.
(ii) We tackle two leakage problems (\ie, data leakage and class leakage) that are most likely to happen during conversion.
(iii) We propose \textit{PrivateSNN} which successfully converts ANNs to SNNs without exposing sensitive information of data.  We encrypt the weight parameters of the converted SNN with a temporal learning rule. Also, distillation from ANN to SNN enables stable encryption training with a small number of training samples.
(iv) We conduct extensive experiments on various datasets including CIFAR10, CIFAR100, and TinyImageNet and demonstrate the advantages of \textit{PrivateSNN} for privacy and energy-efficiency.

 \vspace{-1mm}

\section{Related Work}
\vspace{-1mm}

\subsection{Spiking Neural Networks}
\vspace{-1mm}

Various SNN training techniques have been proposed in order to build efficient neuromorphic systems.
Surrogate gradient learning techniques circumvent the non-differentiable problem of a Leak-Integrate-and-Fire (LIF) neuron by defining an approximate backward gradient function  \cite{lee2016training,lee2020enabling,neftci2019surrogate}.
ANN-SNN conversion techniques convert pre-trained ANNs to SNNs using weight or threshold balancing to replace ReLU with LIF activation \cite{sengupta2019going,han2020rmp,diehl2015fast,rueckauer2017conversion,li2021free}.
Compared to other methods, conversion yields SNNs with competitive accuracy as their ANN counterparts on a variety of tasks.

\begin{figure}[t]
     \centering
         \includegraphics[width=0.4\textwidth]{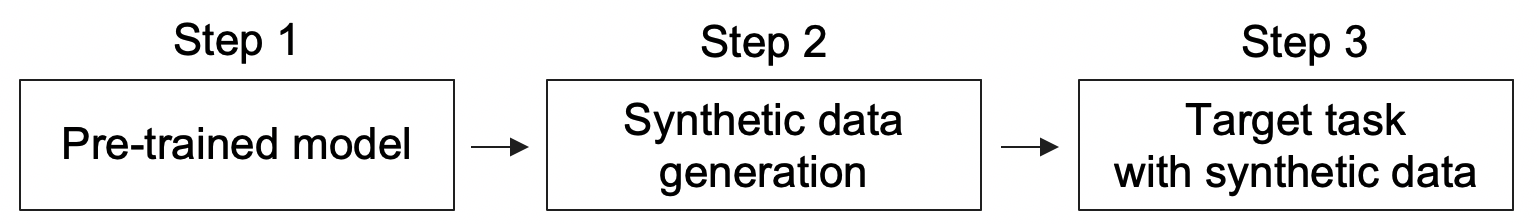}
% \vspace{1mm}
%  \vspace{-3mm}
 \caption{
General approach for addressing data leakage. 
 }
%  \vspace{-2mm}
     \label{fig:intro:dataleak_general}
\end{figure}

\begin{table}[t]
% \addtolength{\tabcolsep}{0.5pt}
% \fontsize{<10pt>}
% \resizebox{0.45\textwidth}{!}{%
\small
\begin{tabular}{lcc}
\toprule
Method &  Data   &  Class   \\
(\cmark: addressed / \xmark: not addressed)   &  Leakage   &  Leakage    \\
\midrule
    Private ANN approaches     &  \cmark & \xmark  \\
    \cite{haroush2020knowledge,nayak2019zero} & & \\
    \hline
    ANN-SNN conversion   &  \xmark   & \xmark    \\
    \cite{sengupta2019going, han2020rmp} & &\\
    \hline
    SNN-surrogate gradients    &  \xmark   & \cmark    \\
    \cite{lee2020enabling,zheng2020going} & &\\
    \hline
    PrivateSNN (ours)  &  \cmark   & \cmark    \\
\bottomrule
\end{tabular}%
% }
%  \vspace{-3mm}
\caption{Related work comparison.} 
\label{table:exp:relatedwork_comparison}
\vspace{-5mm}
\end{table}

\vspace{-2.3mm}
\subsection{{Privacy-preserving Methods}}
\vspace{-1.2mm}

{
Our approach for addressing data leakage is similar to previous privacy-preserving approaches in ANN domain \cite{haroush2020knowledge,nayak2019zero}. Most prior ANN works generate synthetic data from a pre-trained model and then train a model for a target task, as shown in Fig. \ref{fig:intro:dataleak_general}. We conduct ANN-SNN conversion and show successful conversion performance with synthetic data. Note, no prior work before us has shown that SNNs can be converted without using the real dataset. Further, to the best of our knowledge, previous ANN or SNN approaches have not addressed the class leakage issue.
To clarify the objective of our work, in Table
 \ref{table:exp:relatedwork_comparison},  we compare the position of our work with privacy approaches in ANN domain as well as other SNN optimization methods.
The previous privacy preserving ANN approaches successfully address data leakage but cannot resolve class leakage, since ANNs can calculate the exact gradients for reconstructing conceptual class images.
The standard SNN optimization methods (\ie,  ANN-SNN conversion and surrogate gradients learning) cannot address data leakage problem. 
Moreover, ANN-SNN conversion cannot address class leakage since SNNs have the same weights as ANNs, therefore the attacker can recover the original ANN model and perform concept reconstruction.
On the other hand, surrogate gradients learning can prevent the class leakage problem since the weight parameters are trained with non-differentiable LIF activation function. 
Different from the previous methods, our PrivateSNN addresses both problems in a framework, by using data-free conversion and  temporal spike-based training.
}

\vspace{-4mm}

\section{The Vulnerability of ANN-SNN Conversion}

In this section, we present the ANN-SNN conversion algorithm and show the two possible leakage problems.

\vspace{-1mm}
\subsection{ANN-SNN Conversion}
\vspace{-1mm}

Our model is based on LIF neuron.
We formulate the membrane potential $u_{i}^{t}$ of a single neuron $i$ as:
\begin{equation}
    u_i^t = \lambda u_i^{t-1} + \sum_j w_{ij}o^t_j,
    \label{eq:LIF}
\end{equation}
where, $\lambda$ is a leak factor, $w_{ij}$ is a the weight of the connection between pre-synaptic neuron $j$ and post-synaptic neuron $i$.
If the membrane potential $u_i^{t}$ exceeds a firing threshold $\theta$, the neuron $i$ generates spikes $o_i^{t}$.
% which can be formulated as:
% \vspace{-1.5mm}
% \begin{equation}
%     o^{t}_i =
% \begin{cases}
%  1,          & \text{if $u_i^{t}>\theta$},  \\
%     0
%     & \text{otherwise.} 
% \end{cases}
% \label{eq:firing}
% \end{equation}
After the neuron fires, we perform a soft reset, where the membrane potential value $u_i^t$ is lowered by the threshold $\theta$.

We use the method described by \cite{sengupta2019going} for implementing the ANN-SNN conversion. They normalize the weights or the firing threshold $\theta$  to take into account the actual SNN operation in the conversion process. The overall algorithm for the conversion method is detailed in Supplementary D. First, we copy the weight parameters of a pre-trained ANN to an SNN. Then, for every layer, we compute the maximum activation across all time-steps and set the firing threshold to the maximum activation value. The conversion process starts from the first layer and sequentially goes through deeper layers. Note that we do not use batch normalization \cite{ioffe2015batch} since all input spikes have zero mean values. Also, following the previous works \cite{han2020rmp,sengupta2019going,diehl2015fast}, we use Dropout \cite{srivastava2014dropout} for both ANNs and SNNs.

\vspace{-1mm}

\subsection{Two Types of Leakage Problems}
\label{ssection:Two Types of Leakage Problems}

\textbf{Data Leakage from Public Datasets:}
The entire training data is required to convert ANNs to SNNs (see Supplementary D).
However, due to privacy issues, we might not be able to access the training samples. 
Even techniques such as, quantization, model compression, distillation, and domain adaptation have been shown to use synthetically generated data to deploy the final model to preserve privacy  \cite{kundu2020universal,kim2020towards,nayak2019zero,li2021free,haroush2020knowledge,liang2020we}.
Therefore, there is a need to develop a data-free conversion algorithm in the neuromorphic domain.

\begin{algorithm}[t]\small
        \caption{Class representative image generation}
    %   \hspace*{\algorithmicindent} 
       \textbf{Input}: target class ($c$); max iteration ($N$); blurring frequency ($f_{blur}$); learning rate ($\eta$)   \\
      \textbf{Output}:  class representative image $x$
      \begin{algorithmic}[1]
        \State {$x$ $\leftarrow$ $U(0,1)$} \Comment{Initialize input as uniform random distribution}
        \For{$n \gets 1$ to $N$}
            \If {$n$ \% $f_{blur} == 0$}
                \State{$x \leftarrow 
                GaussianBlur(x)$}
            \EndIf
            \State {$y$ $\leftarrow$ $network(x)$} \Comment{compute pre-softmax output $y$}
            \State {$x$ $\leftarrow$ $x + \eta \frac{\partial y_c}{\partial x}$}
        \EndFor
      \end{algorithmic}
          \label{algorithm: overall}
\end{algorithm}

\begin{figure*}[t!]
  \begin{center}
    \includegraphics[width=0.75\textwidth]{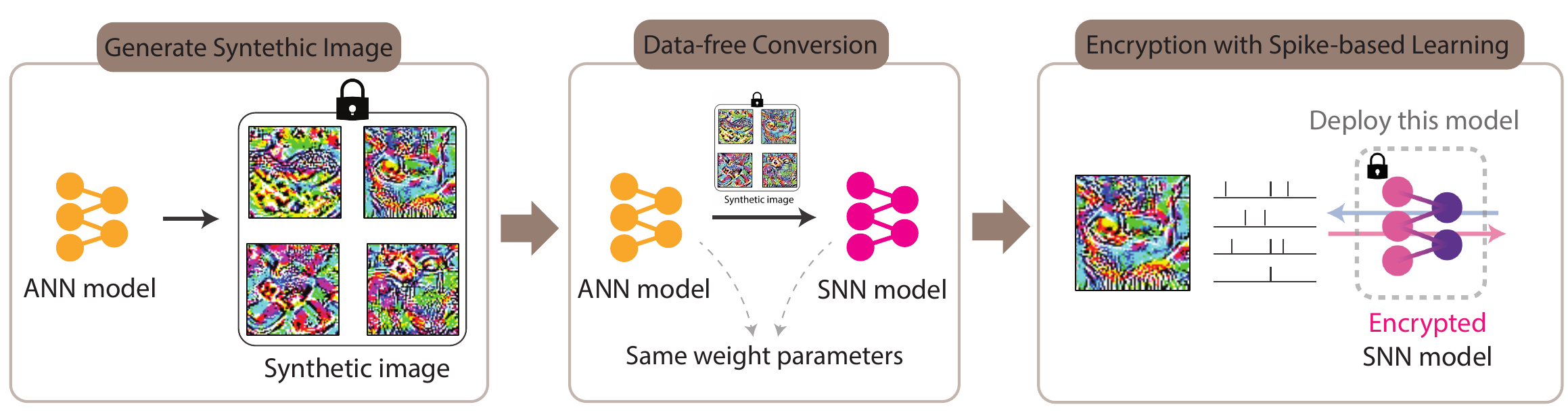}
  \end{center}
    %   \vspace{-4mm}
  \caption{Overview of the proposed \textit{PrivateSNN}. We first generate synthetic images based on underlying data distribution from a pre-trained ANN.
  Then, based on the generated samples, we convert the ANN to a SNN model. 
  Finally, we encode synthetic images to spike signal, and train the converted SNN.
  The weight parameter is encrypted with temporal information. We deploy the final encrypted SNN model for inference.
  }
   \label{fig:method:main_pipeline}
     \vspace{-3mm}
\end{figure*}

\textbf{Class Leakage from Reconverting SNNs to ANNs:}
In addition to data leakage, class information, \eg, pattern and shape of the object,  also can be targeted by the attacker.
Algorithm 1 presents a simple way to obtain class representative information from an ANN model.  
First, we initialize the input tensor with uniform random distribution.
After that, we use an iterative optimization strategy where input noise is updated to maximize the pre-softmax logit $y_{c}$ of target class $c$.
For every blur period $f_{blur}$, we smooth the image with Gaussian blur kernel since gradient at input layer has a high frequency \cite{yosinski2015understanding}.
Fig. \ref{fig:intro:classleak} shows examples of class representative images generated from a pre-trained model.

However, this technique is based on the assumption that we can compute the exact gradient for all layers.
It is difficult to compute gradient value of SNNs due to the non-differentiable nature of LIF neuron (Eq. \ref{eq:LIF}).
Therefore, in order to generate proper class representation, the attacker should reconvert SNNs to ANNs.
The re-conversion process depends on the type of conversion technique; weight scaling or threshold scaling.
There are several conversion algorithms \cite{rueckauer2017conversion,diehl2015fast} that scale weight parameters of each layer.
In such cases, the attacker cannot directly recover original ANN weights.
However, each layer is scaled by a constant value, therefore the original ANN weights might be recovered by searching several combinations of layer-wise scaling factors.
Recent state-of-the-art conversion algorithms \cite{sengupta2019going,han2020rmp,han2020deep} use threshold scaling, \ie,  change the thresholds while maintaining the weight parameters to obtain high performance.
In our experiments, we use threshold scaling for ANN-SNN conversion and then, explore the class leakage issues.
In this case, the original ANN can be reconverted by simply changing LIF neuron to ReLU neuron.
With the reconverted ANN, the attacker can simply reconstruct class representation by backpropagation as shown in Algorithm 1.
Overall, a non-linear weight encryption technique is required to make SNNs robust to class leakage.

\section{Methodology}

This section presents a detailed methodology for \textit{PrivateSNN}.
We first propose a data-free conversion method from a pre-trained ANN.
Then, we describe how a temporal spike-based learning rule can encrypt weight parameters in SNNs. Fig. \ref{fig:method:main_pipeline} illustrates the overall approach.

\subsection{Data-Free Conversion for Data Leakage}

\textbf{Data Generation from a Pre-trained ANN:}
Without accessing real data, we generate synthetic images from a pre-trained ANN. 
Conversion performance relies on the maximum activation value of features, therefore synthetic images have to carefully  reflect underlying data distribution from the pre-trained ANN.
\citet{nayak2019zero} take into account the relationship between classes in order to generate data, resulting in better performance on a distillation task.  
Following this pioneering work, we generate synthetic images based on class relationships from the weights of the last fully-connected layer.
Specifically, we can define a weight vector $w_{c}$ between the penultimate layer and the class logit $c$ in the last layer. Then, we calculate the class similarity score between class $i$ and $j$:
\begin{equation}
    s_{ij} = \frac{w_i^T w_j}{\|w_i\|_2^2 \|w_j\|_2^2}.
\end{equation}
Then, we sample a soft label based on Dirichlet distribution where a concentration parameter $\alpha_c$ is a class similarity vector of class $c$.
For each class $c$, the class similarity vector consists of class similarity between class $c$ and other classes.
For example, for class $1$ of CIFAR10 dataset, the concentration parameter is $\alpha_1 = [s_{10}, s_{11}, s_{12}, ..., s_{19}]$. 
For a sampled soft label from Dirichlet distribution, we optimize input $x$ initialized with uniform random distribution.
We collect the same number of samples for each class and exploit this synthetic dataset for conversion. The overall procedure for data-free conversion is shown in Algorithm 2 [Step1].

\begin{algorithm}[t]\small
        \caption{PrivateSNN Approach}
    %   \hspace*{\algorithmicindent} 
       \textbf{Input}:  total class set ($C$); the number of samples per class ($N$);  pre-trained ANN model ($ANN$); SNN model ($SNN$); spike time-step ($T$); distillation temperature ($\tau$); \\
      \textbf{Output}:  encrypted SNN model
      \begin{algorithmic}[1]
        % \State{\textbf{begin}}
        %
        \State {\textbf{[Step1] Data-free conversion}}
        \State {$D$ $\leftarrow$ $\O$}\Comment{Initialize synthetic dataset $D$}
        \For{$c \gets 1$ to $C$}
                 \State {$\alpha_c$ $\leftarrow$ $ANN.fc.weight$}\Comment{Compute class similarity}
            \For{$n \gets 1$ to $N$}
                \State {$y \sim Dir(\alpha_c)$} \Comment{Sample soft label from Dirichlet}
                 \State {$x$ $\leftarrow$ $U(0,1)$} \Comment{Initialize input}
                \State {Find $x$ that minimizes $L_{CE}(ANN(x), y)$}
                \State{$D \gets D \cup \{x\}$}
            \EndFor
        \EndFor
        \State{Do ANN-SNN conversion  with synthetic data $D$}
        % \vspace{1mm}
        \State {\textbf{[Step2] Encryption with spike-based learning}}
        \For{$i \gets 1$ to $max\_iter$}
        \State {fetch a mini batch $X \subset D$}
        \For{$t \gets 1$ to $T$}  
            \State{O $\leftarrow$ PoissonGenerator(X)}
            \For{$l \gets 1$ to $L-1$} 
                \State{$(O^t_{l}, U_l^{t}) \leftarrow (\lambda, U_l^{t-1}, (W_{l}, O^{t}_{l-1}))$}
                \Comment{Eq. \ref{eq:LIF}}
            \EndFor
            \State{$U_L^{t} \hspace{-1mm} \leftarrow  \hspace{-1mm} ( U_L^{t-1},  (W_{l}, O^{t}_{L-1}))$ } \Comment{ Final layer}
        \EndFor
        \State{$L_{CE} \leftarrow (U_L^T, Y)$} 
        \State{$L_{KD} \leftarrow (ANN(X,\tau), SNN(X,\tau))$} 
        \Comment{Eq. \ref{eq:total_loss}}
        \State{Do back-propagation and weight update}
    \EndFor
      \end{algorithmic}
          \label{algorithm: overall}
\end{algorithm}

\textbf{ANN-SNN Conversion with synthetic images:}
By generating synthetic images, we do not have to access the original dataset. 
Instead, we find the threshold of each layer from the maximum activation of synthetic images (see ANN-SNN Conversion algorithm in Supplementary D).
Interestingly, we observe that converted SNNs with synthetic images almost recover the performance of the original ANN. This means that conversion process is feasible with inherent data distribution from a trained model.
However, SNNs are still vulnerable to class leakage.
If the attacker gets access to the weights of the SNN, they can easily recover the original ANN by simply changing the LIF neuron to ReLU.
Therefore, we encrypt the converted SNN using the temporal spike-based learning rule (will be described in the next subsection).
Note, we use a relatively small number of time-steps (\eg, 100 $\sim$ 150) for conversion.
This is because short time-steps reduces training time and memory for post-conversion training. Also, short latency can bring more energy efficiency at inference.  
We observe that encryption training recovers the performance loss caused by using small number of time-steps for conversion.

\vspace{-1mm}

\subsection{Class Encryption with Spike-based Training}
\vspace{-1mm}

The key idea here is that training SNNs with temporal data representation makes class information difficult to be interpreted in the spatial domain.
\vspace{-1mm}

\textbf{Encoding static images with rate coding}:
In order to map static images to temporal signals, we use rate coding (or Poisson coding). 
Given the time window, rate coding generates spikes where the number of spikes is proportional to the pixel intensity.
For time-step $t$, we generate a random number for each pixel $(i,j)$  with normal distribution ranging between $[I_{min}, I_{max}]$, where $I_{min}, I_{max}$ correspond to the minimum and  maximum possible pixel intensity.
After that, for each pixel location, we compare the pixel intensity with the generated random number.
If the random number is greater than the pixel intensity, the Poisson spike generator outputs a spike with amplitude $1$.
Otherwise, the Poisson spike generator does not yield any spikes. 
Overall, rate coding enables static images to span the temporal axis without huge information loss.

\textbf{Training SNNs with spike-based learning}:
Given input spikes, we train converted SNNs based on gradient optimization. 
Intermediate LIF neurons accumulate pre-synaptic spikes and generate output spikes (Eq. \ref{eq:LIF}).
Spike information is passed through all layers and stacked or accumulated at the output layer (\ie, prediction layer)
This enables the accumulated temporal spikes to be represented as probability distribution after the softmax function.
From the accumulated membrane potential, we can define the cross-entropy loss for SNNs.
% as:
% \begin{equation}
%     {L_{CE}} = - \sum_{i} y_{i} log(\frac{e^{u_i^T}}{\sum_{k=1}^{C}e^{u_k^T}}),
%     \label{eq:celoss}
% \end{equation}
% where, $y$ and $T$ stand for the ground-truth label and the total number of time-steps, respectively.

\textbf{On-device distillation}: One important thing we have to consider is the computational cost for post-conversion training.
This is crucial to a limited-resource environment such as a mobile device with battery constraints.
Also, SNNs require multiple feed-forward steps per one image, and thus are more energy-consuming compared to ANN training.
In order to reduce the cost for training, we can reduce the number of synthetic training samples.
However, with a small number of training samples, the networks  easily overfit, resulting in performance degradation.
To address this issue, we distill knowledge from ANNs to SNNs.
 \citet{yuan2019revisit} recently discovered the connection between knowledge distillation and label smoothing,  which supports distillation improves the generalization power of the model.
 Thus, it is intuitive that if we use knowledge distillation during the spike-based training of the converted SNN, then the model is likely to show better generalization to small number of data samples.
Therefore, the total loss function becomes the combination of cross-entropy loss and distillation loss:
\begin{equation}
    {L} = (1-m) L_{CE} + m L_{KD}(A(X,\tau), S(X,\tau)).
    \label{eq:total_loss}
\end{equation}
Here, $A(\cdot)$ and $S(\cdot)$ represent ANN and SNN models, respectively. Also, $\tau$ denotes distillation temperature, and $m$ is the balancing coefficient between two losses.
Note that $L_{KD}$ is knowledge distillation loss \cite{hinton2015distilling}.
The training samples $X$ used in this stage are a subset of the synthetically generated data used during conversion.

Based on Eq. \ref{eq:total_loss}, we compute the gradients of each layer $l$.
Here, we use spatio-temporal back-propagation (STBP), which accumulates the gradients over all time-steps \cite{wu2018spatio,neftci2019surrogate}.
We can formulate the gradients at the layer $l$ by chain rule as:
\begin{equation}
      \frac{\partial L}{\partial W_l} =
\begin{cases}
 \sum_{t}(\frac{\partial L}{\partial O_l^t}\frac{\partial O_l^t}{\partial U_l^t} + \frac{\partial L}{\partial U_l^{t+1}}  \frac{\partial U_l^{t+1}}{\partial U_l^{t}})
 \frac{\partial U_l^t}{\partial W_l},  & \text{if $l:$ hidden } \\
    \frac{\partial L}{\partial U_l^T}\frac{\partial U_l^T}{\partial W_l}.
    & \text{if $l:$ output} 
\end{cases}
\label{eq:delta_W}
\end{equation}
Here, $O^t_l$ and $U^t_l$ are output spikes and membrane potential at time-step $t$ for layer $l$, respectively.
For the output layer, we get the derivative of the loss $L$ with respect to the membrane potential $u_i^T$ at final time-step $T$, which
%
% \begin{equation}
% \frac{\partial L}{\partial u_i^T} =\frac{e^{u_i^T}}{\sum_{k=1}^{C}e^{u_k^T}} - y_i.
% \end{equation}
is continuous and differentiable for all possible membrane potential values.
On the other hand, LIF neurons in hidden layers  generate  spike output only if the membrane potential $u_i^t$ exceeds the firing threshold, leading to non-differentiability.
To deal with this problem, we introduce an approximate gradient:
\begin{equation}
    \frac{\partial o_i^t}{\partial u_i^t} = \max \{0, 1-  \ | \frac{u_i^t - \theta}{\theta} \ | \}.
    \label{eq:approx_grad_function}
\end{equation}
Overall, we update the network parameters at the layer $l$ based on the gradient value (Eq. \ref{eq:delta_W}) as $W_l = W_l - \eta \Delta W_l$. The procedure for spike-based training for encrypting the model is detailed in Algorithm 2 [Step2].

\section{Attack Scenarios for Class Leakage}
\label{section:two_attack_scenarios}
\vspace{-0.5mm}
In this section, we present two possible attack scenarios on class leakage. Since we do not access the original training data, the data leakage problem is  addressed. Therefore, here we only discuss the class leakage problem. 

\textbf{Attack scenario 1 (Reconverting SNN to ANN):}
As discussed earlier, the attacker might copy the weights of SNN and recover the original ANN.
By using the re-converted ANN, the attacker optimizes the input noise based on Algorithm 1.
Without post-conversion encryption training, the attacker simply reconstructs class representation by backpropagation. 
However, if we use spike-based training, the weight of SNN is fully encrypted in the spatial domain.

\textbf{Attack scenario 2 (Directly generate class representation from SNN):}
The attacker might directly backpropagate gradients in SNN architecture and reconstruct class representation.
Thus, this is the SNN version of Algorithm 1.
The technical problem here is that LIF neurons and  Poisson spike generation process are non-differentiable.
To address this issue, we use approximated gradient functions (Eq. \ref{eq:approx_grad_function}) for LIF neurons.
Also, in order to convert the gradient in the temporal domain to the spatial domain, we accumulate gradient at the first convolution layer.
After that, we deconvolve the accumulated gradients with weights of the first layer. 
These deconvolved gradients have a similar value with original gradients of images before the Poisson spike generator, which has been validated in previous work \cite{sharmin2020inherent}.
Thus, we can get a gradient $\delta x$ converted into spatial domain, and the input noise is updated with gradient $\delta x$ scaled by $\zeta$.
Algorithm 3 illustrates the overall optimization process.
However, this attack does not show meaningful features due to the discrepancy between real gradients and approximated gradients. 
This supports that SNN model itself is robust to gradient-based security attacks \cite{roy2019towards,sharmin2020inherent}.
We show the qualitative results for Attack 1, 2 scenarios in Fig. \ref{fig:exp:qualitative_timg}.

\begin{algorithm}[t]\small
        \caption{Directly generate class representation from SNNs (Attack scenario 2)}
    %   \hspace*{\algorithmicindent} 
       \textbf{Input}: target class ($c$); max iteration ($N$);  scaling factor ($\zeta$); SNN  model (SNN); spike time-step ($T$) \\
      \textbf{Output}:  class representative image $x$
      \begin{algorithmic}[1]
        \State {$x$ $\leftarrow$ $U(0,1)$} \Comment{Initialize input}
        \State {$W_1$ $\leftarrow$ $SNN.conv1.weight$}\Comment{ First convolution weight}
        \State {$G = 0$}\Comment{Accumulated gradient at the first conv layer}
        \For{$n \gets 1$ to $N$}
            \For{$t \gets 1$ to $T$}
            \State{$y \gets SNN(x_t)$}
            \State{$G += \frac{1}{T}\frac{\partial y_c}{\partial x_{conv1}}$} \Comment{Accumulate gradients in layer 1}
            \EndFor
            \State {$\delta x \gets Deconvolution(W_1, G)$} \Comment{Deconvolution}
            \State {$x$ $\leftarrow$ $x + \zeta \delta x$}
        \EndFor
      \end{algorithmic}
          \label{algorithm: overall}
\end{algorithm}

\section{Experiments}
\vspace{-0.5mm}

\subsection{Experimental Setting}
\vspace{-1mm}
We evaluate our \textit{PrivateSNN} on three public datasets (\ie, CIFAR-10 \cite{krizhevsky2009learning}, CIFAR-100 \cite{krizhevsky2009learning}, Tiny-ImageNet \cite{deng2009imagenet}).
% \textbf{CIFAR-10} \cite{krizhevsky2009learning} consists of 60,000 images (50,000 for training / 10,000 for testing) with 10 categories. All images are RGB color images whose size are 32 $\times$ 32.
% \textbf{CIFAR-100} has the same configuration as CIFAR-10, except it contains images from 100 categories.
% \textbf{Tiny-ImageNet} is the modified subset of the original ImageNet dataset.
% Here, there are 200 different classes of ImageNet dataset \cite{deng2009imagenet}, with 100,000 training and 10,000
% validation images. The resolution of the
% images is 64$\times$64 pixels.
Our implementation is based on Pytorch \cite{paszke2017automatic}. 
For conversion, we apply threshold scaling technique as proposed in \cite{han2020rmp}.
For post-conversion training, we use Adam with base learning rate 1e-4.
Here, we use 5000, 10000, 10000 synthetic samples for training SNNs on CIFAR10, CIFAR100, and TinyImageNet, respectively.
We use step-wise learning rate scheduling with a decay factor of 10 at 50\% and 70\% of the total number of epochs. We set the total number of epochs to 20 for all datasets.
For on-device distillation, we set $m$ and  $\tau$ to $0.7$ and $20$ in Eq. \ref{eq:total_loss}, respectively. 
For class representation, we set $f_{blur}$ and $\eta$ in Algorithm 1 to 4 and 6, respectively.
For attack scenario 2 in Algorithm 3, we set $\zeta$ to 0.01.
All detailed experimental setup and hyperparameters are described in Supplementary B.
Note that our objective is to showcase the advantages of \textit{PrivateSNN} for tackling data and class leakage.

\vspace{-1mm}

\subsection{Performance Comparison}
\vspace{-1mm}
{
Before we present the experimental results, we define the terms used in our method: 
\textit{Data-free Conversion (DC)}, 
\textit{Class Encryption Training (CET)}, and 
\textit{On-device Distillation (OD)}.
We call our final method as \textit{PrivateSNN}, thus, \textit{PrivateSNN = DC + CET + OD}.
}
 
Surprisingly, we find that \textit{PrivateSNN} encrypts the networks without significant performance loss.
Table \ref{table:exp:performance} shows the performance of reference ANN (\ie, VGG16) and previous conversion methods which uses training data.
Here, we use \citet{sengupta2019going,han2020rmp,zambrano2019sparse} as representative state-of-the-art conversion methods for comparison.
The results show that our \textit{PrivateSNN} can be designed without a huge performance drop across all datasets.
This implies that synthetic samples generated from the ANN model contain enough information for a successful conversion and post-training processes.
 
In Table \ref{table:exp:ablation}, we conduct ablation studies on each component (\ie, DC, CET, and OD) of our method. Here, we present the robustness of model on data leakage and class leakage and compare the number of training samples, and classification accuracy. 
With DC, the SNN model can address data leakage, however, it is still vulnerable to class leakage.
Adding CET on DC resolves class leakage with performance improvement.
However, an insufficient number of training samples cannot achieve near state-of-the-art performance.
Finally, using OD helps the networks to improve the performance with a limited number of samples.

\begin{table}[t]
\small
% \addtolength{\tabcolsep}{0.5pt}
\centering
% \resizebox{0.5\textwidth}{!}{%
\begin{tabular}{lccc}
\toprule
Method &  Require  & Dataset  &     Acc (\%) \\
       &   train data?  &   &    \\
\midrule
    VGG16  & -  & CIFAR10 & 91.6  \\
    \citet{rueckauer2017conversion} &  Yes  & CIFAR10  &  90.9  \\
    \citet{sengupta2019going} &  Yes  & CIFAR10  &  91.5  \\
    \citet{han2020rmp} &  Yes  & CIFAR10  &  91.4 \\
    \citet{zambrano2019sparse} &  Yes  & CIFAR10  &  89.7 \\
    PrivateSNN (ours) &  No  & CIFAR10  &  89.2 \\
    %%%%%%%%%%%%%%%%%%%%%%%%%%
    \midrule
    VGG16  & -  & CIFAR100 & 64.3  \\
    \citet{sengupta2019going} &  Yes  & CIFAR100  &  62.7 \\
     \citet{zambrano2019sparse} &  Yes  & CIFAR100  &  63.4  \\
    PrivateSNN (ours) &  No  & CIFAR100  &  62.3\\
        %%%%%%%%%%%%%%%%%%%%%%%%%%
     \midrule
    VGG16  & -  & TinyImgNet & 51.9  \\
    \citet{sengupta2019going} &  Yes  & TinyImgNet  &  50.6 \\
     PrivateSNN (ours) &  No  & TinyImgNet  &  50.7 \\
\bottomrule
\end{tabular}%
% }
% \vspace{-3mm}
\caption{Classification Accuracy (\%) on  CIFAR10, CIFAR100, and TinyImageNet.  We report the accuracy of VGG16 (pre-trained ANN) \cite{simonyan2014very} architecture in our experiments as a reference. 
}
\label{table:exp:performance}
  \vspace{-1mm}
\end{table}

\begin{table}[t]
\small
% \addtolength{\tabcolsep}{0.5pt}
\centering
% \resizebox{0.47\textwidth}{!}{%
\begin{tabular}{lcccc}
\toprule
Method &  D.L.   & C.L. & \# Train data  &    Acc (\%) \\
\midrule
    DC (T=150)   &  \cmark  & \xmark  & - &  82.8  \\
    DC + CET &  \cmark  & \cmark & 5000 &  86.9 \\
    DC + CET + OD &  \cmark  & \cmark & 5000  &  89.2 \\
    %%%%%%%%%%%%%%%%%%%%%%%%%%
\bottomrule
\end{tabular}%
% }
% \vspace{-3mm}
\caption{Ablation study for each component in our method on CIFAR10 dataset (\cmark: addressed / \xmark: not addressed).  D.L / C.L. denote data leakage and class leakage, respectively.}
\label{table:exp:ablation}
  \vspace{-4mm}
\end{table}

\begin{figure}[t]
\begin{center}
\def\arraystretch{0.5}
\begin{tabular}{@{}c@{}c@{}c}
\includegraphics[width=0.38\linewidth]{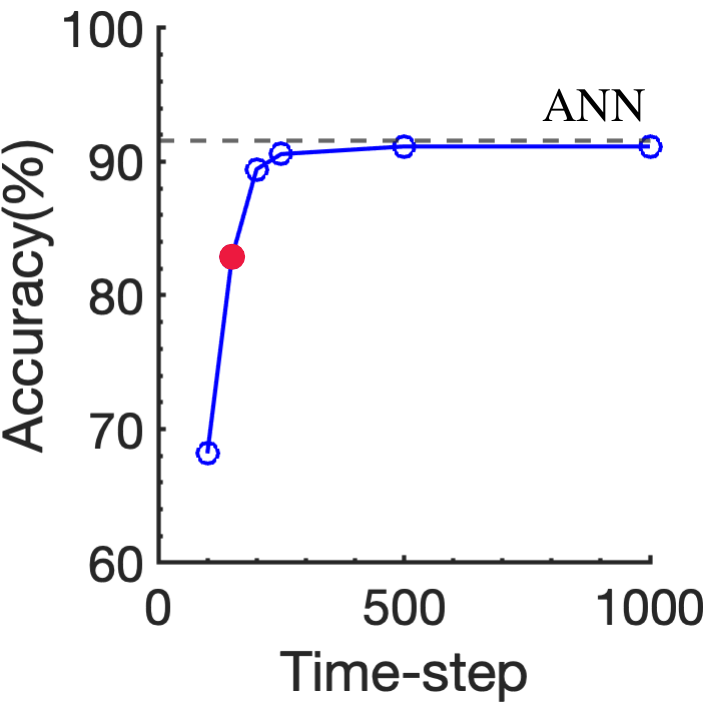} &
\includegraphics[width=0.64\linewidth]{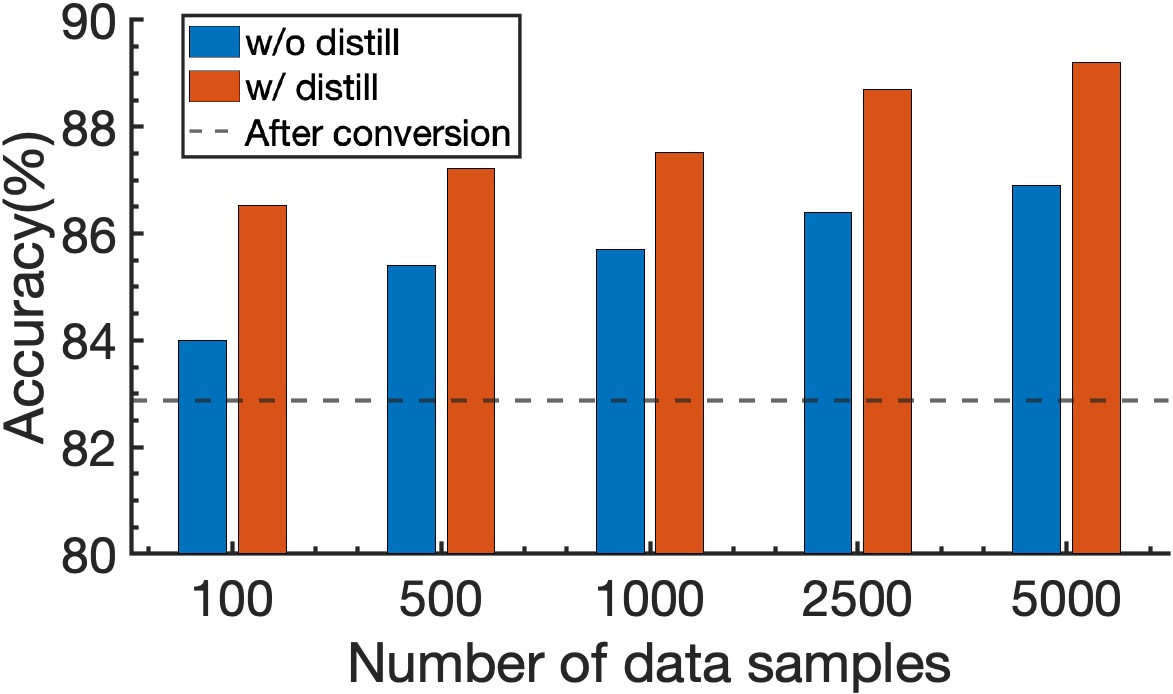}&
\\
{\hspace{5mm}(a)  } & {\hspace{3mm}(b)  }\\
\end{tabular}
% \vspace*{-0.1in}
%   \vspace{-5mm}
\end{center}
\caption{ 
(a) Data-free Conversion performance. 
We use a small number of time-steps for efficient post-conversion training (marked by red dot).
(b) Performance with respect to the number of training samples using encryption training. 
For both experiments, we use VGG16 trained on CIFAR10.
}
%   \vspace{-1mm}
\label{fig:exp:time_numsample_ablation}
\end{figure}

\vspace{-1mm}

\subsection{Analysis on Data-free Conversion}
\vspace{-1mm}

In Fig. \ref{fig:exp:time_numsample_ablation}(a), we measure the data-free conversion performance with respect to the number of time-steps in the conversion process.
The results show that the data-free SNN conversion model can almost recover the original ANN performance with large number of time-steps (\ie, $\ge 500$). But, we use a small number of time-steps (\ie, $\le 200$) to perform the conversion and then, perform CET in the low time-step regime.
Here, we only show CIFAR10 results. The CIFAR100, and TinyImageNet results are shown in Supplementary E.
Specifically, \textit{PrivateSNN} is trained with time-step 150, 200, and 200 for CIFAR10, CIFAR100, and TinyImageNet,
This is because smaller time-steps bring more energy-efficiency during both training and inference.
Also, we observe that encryption training with distillation recovers the accuracy even though SNNs are converted in a low time-step regime (Table \ref{table:exp:ablation}).

\vspace{-1mm}

\subsection{Effect of Distillation in Encryption Training}
\vspace{-1mm}

After converting ANNs to SNNs, we train the networks with synthetic images.
In order to figure out how many samples are required for post training, we show the performance with respect to the number of synthetic samples in Fig. \ref{fig:exp:time_numsample_ablation}(b).
Note, data-free conversion at 150 time-steps achieves 82.8 \% accuracy (black dotted line in the figure).
The results show that encryption training (without distillation) degrades the performance with a small number of samples.
Interestingly, with distillation, the network almost preserves the performance regardless of sample size.
Thus,  distillation is an effective regularization method for spike-based learning with a small number of synthetic images.

\begin{table}[t]
% \addtolength{\tabcolsep}{0.5pt}
\small
\centering
% \vspace{-3mm}
\resizebox{0.48\textwidth}{!}{%
\begin{tabular}{lcc}
\toprule
Method &  w/o Encryption  &   w/ Encryption  \\
\midrule
    Attack scenario 1  &  354.8 & 448.2  \\
     Attack scenario 2 &  447.9 & 422.4  \\
\bottomrule
\end{tabular}%
}
% \vspace{-2.5mm}
\caption{FIDs between test set and generated images from reconverted ANN ({Attack scenario 1}) and backward gradients on SNNs ({Attack scenario 2}) on CIFAR10. 
Higher FID implies lower leakage. 
}
\label{table:exp:FID}
\vspace{-4mm}
\end{table}

\vspace{-1mm}

\subsection{Robustness on Class Leakage Problem}
\begin{figure}[t]
     \centering
         \includegraphics[width=0.40\textwidth]{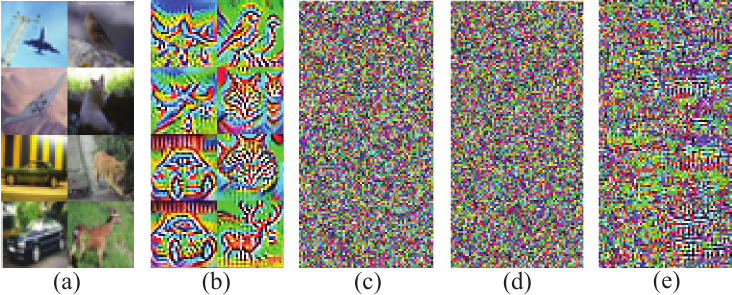}
% \vspace{1mm}
%  \vspace{-3mm}
 \caption{
Qualitative results (CIFAR10) of two attack scenarios on the class leakage problem.
{(a) Original Data. } 
{(b) Attack1-w/o CET. }
{(c) Attack1-w/ CET.}
{(d) Attack2-w/o CET. }
{(e) Attack2-w/ CET. }
 }
%  \vspace{-1mm}
     \label{fig:exp:qualitative_timg}
\end{figure}

\vspace{-1mm}
To validate the robustness of \textit{PrivateSNN} against Attack scenario 1 and Attack scenario 2 (two most likely class leakage scenarios), we synthesize the class representation of SNNs with (w/) and without (w/o) class encryption training (CET).
Fig. \ref{fig:exp:qualitative_timg} shows 8 examples of generated images from CIFAR10 (The results from CIFAR100, and TinyImageNet are shown in Supplementary H). 
We visualize images from five configurations: original images, Attack1 w/o CET, Attack1 w/ CET, Attack2 w/o CET, and Attack2 w/ CET.
We observe that SNNs without CET (\ie,  simply, data free converted SNNs) are vulnerable to Attack1, showing important features of original classes (see Fig. \ref{fig:exp:qualitative_timg}(b)). On the other hand, with temporal spike-based learning rule, 
Attack1 does not discover any meaningful information as shown in Fig. \ref{fig:exp:qualitative_timg}(c). 
For Attack2 (Fig. \ref{fig:exp:qualitative_timg}(d) and Fig. \ref{fig:exp:qualitative_timg}(e)), synthetic images show noisy results due to discrepancy between real gradients and approximated gradients. This comes from the intrinsic nature of SNNs, therefore SNNs are robust to Attack2 even without encryption.
Overall, \textit{PrivateSNN} is an effective solution to class leakage problem.

We quantify the security of the model by measuring how much generated images represent similar features with that of original images.
To this end, we use \textit{F\`rechet inception distance} (FID)  metric  \cite{heusel2017gans} that is widely used in GAN evaluation \cite{miyato2018cgans,miyato2018spectral}. 
The FID score compares the statistics of embedded features in a feature space 
(see Supplementary C for more detailed explanation).
Thus, a lower FID score means that the generated images provide  the original data-like feature.
In Table \ref{table:exp:FID}, the SNN model without encryption training on attack scenario 1 achieves a much lower FID score (\ie, 354.8) compared to others, which supports our visualization results.

\vspace{-1mm}
\subsection{ Energy-efficiency of PrivateSNN}
\vspace{-1mm}

For the sake of complete analysis, we calculate the inference energy of \textit{PrivateSNN} and compare with ANN and other conversion methods. Note that, we use the same energy estimation model as \cite{panda2020toward}, which is rather a rough estimate that considers only Multiply and Accumulate (MAC) operations and neglects memory and peripheral circuit energy.
We calculate the energy based on spike rate (\ie, the average number of spikes across time) of all layers  (see Supplementary F for details).
In Table \ref{table: energy_consumption}, we compare the energy efficiency between VGG16, standard conversion \cite{sengupta2019going,han2020rmp}, and our method.
The results show that  \textit{PrivateSNN} is more efficient than both ANN as well as a standard converted SNN. This implies our approach of $DC+CET+OD$ lowers the overall spike rate which makes  \textit{PrivateSNN} more energy-efficient.

%****FIGURE 2**************************************
%******************

\begin{table}[t]
\small
\centering
\resizebox{0.47\textwidth}{!}{%
\begin{tabular}{lccc}
\toprule
Method &  Timestep  & Acc (\%) & $\frac{E_{ANN}}{E_{ours}}$ \\
\midrule
    VGG16 (ANN)  &  1   & 91.5 &  1$\times$ \\
    \citet{sengupta2019going} & 500 & 91.2  &  27.9$\times$ \\
    \citet{han2020rmp} & 250 & 91.4  &  53.9$\times$ \\
    PrivateSNN (ours) &  150  & 89.2   &  55.8$\times$ \\
\bottomrule
\end{tabular}%
}
% \vspace{-3mm}
\caption{Energy efficiency comparison on VGG16-CIFAR10.}
\vspace{-4mm}
\label{table: energy_consumption}
\end{table}
%  \vspace{-2mm}

\vspace{-1.5mm}
\section{Conclusion}
\vspace{-1mm}
% \vspace{-2mm}
For the first time, we expose the vulnerability of converted SNNs to data and class leakage. We propose \textit{PrivateSNN} that comprises of data-free conversion followed by weight encryption with spike-based training on synthetic data to tackle the privacy issues. 
We further optimize the training process with distillation that enables stable encryption training with very few training samples. 
So far, the discussion around SNNs has more or less been limited to energy-efficiency. This work sets precedence on the vulnerabilities unique to the SNN domain and also showcases the benefits of temporal spike-based learning for encryption. We hope this fosters future work around security and privacy in SNNs.
One limitation of data-free conversion is that the conversion performance is reduced with the overfitted networks which are likely to generate biased data representation. We discuss such limitation in Supplementary.

\section*{Acknowledgments}
The research was funded in part by C-BRIC, one of six centers in JUMP, a Semiconductor Research Corporation (SRC) program sponsored by DARPA, the National Science Foundation (Grant$\#$1947826), and Amazon Research Award.

\bibliography{ref}

\begin{thebibliography}{48}
\providecommand{\natexlab}[1]{#1}

\bibitem[{Akopyan et~al.(2015)Akopyan, Sawada, Cassidy, Alvarez-Icaza, Arthur,
  Merolla, Imam, Nakamura, Datta, Nam et~al.}]{akopyan2015truenorth}
Akopyan, F.; Sawada, J.; Cassidy, A.; Alvarez-Icaza, R.; Arthur, J.; Merolla,
  P.; Imam, N.; Nakamura, Y.; Datta, P.; Nam, G.-J.; et~al. 2015.
\newblock Truenorth: Design and tool flow of a 65 mw 1 million neuron
  programmable neurosynaptic chip.
\newblock \emph{IEEE transactions on computer-aided design of integrated
  circuits and systems}, 34(10): 1537--1557.

\bibitem[{Christensen et~al.(2022)Christensen, Dittmann, Linares-Barranco,
  Sebastian, Le~Gallo, Redaelli, Slesazeck, Mikolajick, Spiga, Menzel
  et~al.}]{christensen20222022}
Christensen, D.~V.; Dittmann, R.; Linares-Barranco, B.; Sebastian, A.;
  Le~Gallo, M.; Redaelli, A.; Slesazeck, S.; Mikolajick, T.; Spiga, S.; Menzel,
  S.; et~al. 2022.
\newblock 2022 roadmap on neuromorphic computing and engineering.
\newblock \emph{Neuromorphic Computing and Engineering}.

\bibitem[{Davies et~al.(2018)Davies, Srinivasa, Lin, Chinya, Cao, Choday,
  Dimou, Joshi, Imam, Jain et~al.}]{davies2018loihi}
Davies, M.; Srinivasa, N.; Lin, T.-H.; Chinya, G.; Cao, Y.; Choday, S.~H.;
  Dimou, G.; Joshi, P.; Imam, N.; Jain, S.; et~al. 2018.
\newblock Loihi: A neuromorphic manycore processor with on-chip learning.
\newblock \emph{IEEE Micro}, 38(1): 82--99.

\bibitem[{Deng et~al.(2022)Deng, Li, Zhang, and Gu}]{deng2022temporal}
Deng, S.; Li, Y.; Zhang, S.; and Gu, S. 2022.
\newblock Temporal Efficient Training of Spiking Neural Network via Gradient
  Re-weighting.
\newblock \emph{arXiv preprint arXiv:2202.11946}.

\bibitem[{Diehl and Cook(2015)}]{diehl2015unsupervised}
Diehl, P.~U.; and Cook, M. 2015.
\newblock Unsupervised learning of digit recognition using
  spike-timing-dependent plasticity.
\newblock \emph{Frontiers in computational neuroscience}, 9: 99.

\bibitem[{Diehl et~al.(2015)Diehl, Neil, Binas, Cook, Liu, and
  Pfeiffer}]{diehl2015fast}
Diehl, P.~U.; Neil, D.; Binas, J.; Cook, M.; Liu, S.-C.; and Pfeiffer, M. 2015.
\newblock Fast-classifying, high-accuracy spiking deep networks through weight
  and threshold balancing.
\newblock In \emph{2015 International Joint Conference on Neural Networks
  (IJCNN)}, 1--8. ieee.

\bibitem[{Fang et~al.(2019)Fang, Wang, Gomez, Datta, Khan, and
  Raychowdhury}]{fang2019swarm}
Fang, Y.; Wang, Z.; Gomez, J.; Datta, S.; Khan, A.~I.; and Raychowdhury, A.
  2019.
\newblock A swarm optimization solver based on ferroelectric spiking neural
  networks.
\newblock \emph{Frontiers in neuroscience}, 13: 855.

\bibitem[{Frady et~al.(2020)Frady, Orchard, Florey, Imam, Liu, Mishra, Tse,
  Wild, Sommer, and Davies}]{frady2020neuromorphic}
Frady, E.~P.; Orchard, G.; Florey, D.; Imam, N.; Liu, R.; Mishra, J.; Tse, J.;
  Wild, A.; Sommer, F.~T.; and Davies, M. 2020.
\newblock Neuromorphic Nearest Neighbor Search Using Intel's Pohoiki Springs.
\newblock In \emph{Proceedings of the Neuro-inspired Computational Elements
  Workshop}, 1--10.

\bibitem[{Furber et~al.(2014)Furber, Galluppi, Temple, and
  Plana}]{furber2014spinnaker}
Furber, S.~B.; Galluppi, F.; Temple, S.; and Plana, L.~A. 2014.
\newblock The spinnaker project.
\newblock \emph{Proceedings of the IEEE}, 102(5): 652--665.

\bibitem[{Goodfellow, Shlens, and Szegedy(2014)}]{goodfellow2014explaining}
Goodfellow, I.~J.; Shlens, J.; and Szegedy, C. 2014.
\newblock Explaining and harnessing adversarial examples.
\newblock \emph{arXiv preprint arXiv:1412.6572}.

\bibitem[{Han and Roy(2020)}]{han2020deep}
Han, B.; and Roy, K. 2020.
\newblock Deep Spiking Neural Network: Energy Efficiency Through Time based
  Coding.
\newblock In \emph{Proc. IEEE Eur. Conf. Comput. Vis.(ECCV)}, 388--404.

\bibitem[{Han et~al.(2020)}]{han2020rmp}
Han, B.; et~al. 2020.
\newblock RMP-SNN: Residual Membrane Potential Neuron for Enabling Deeper
  High-Accuracy and Low-Latency Spiking Neural Network.
\newblock In \emph{Proceedings of the IEEE/CVF Conference on Computer Vision
  and Pattern Recognition}, 13558--13567.

\bibitem[{Haroush et~al.(2020)Haroush, Hubara, Hoffer, and
  Soudry}]{haroush2020knowledge}
Haroush, M.; Hubara, I.; Hoffer, E.; and Soudry, D. 2020.
\newblock The knowledge within: Methods for data-free model compression.
\newblock In \emph{Proceedings of the IEEE/CVF Conference on Computer Vision
  and Pattern Recognition}, 8494--8502.

\bibitem[{He et~al.(2016)He, Zhang, Ren, and Sun}]{he2016deep}
He, K.; Zhang, X.; Ren, S.; and Sun, J. 2016.
\newblock Deep residual learning for image recognition.
\newblock In \emph{Proceedings of the IEEE conference on computer vision and
  pattern recognition}, 770--778.

\bibitem[{Heusel et~al.(2017)Heusel, Ramsauer, Unterthiner, Nessler, and
  Hochreiter}]{heusel2017gans}
Heusel, M.; Ramsauer, H.; Unterthiner, T.; Nessler, B.; and Hochreiter, S.
  2017.
\newblock Gans trained by a two time-scale update rule converge to a local nash
  equilibrium.
\newblock \emph{arXiv preprint arXiv:1706.08500}.

\bibitem[{Hinton, Vinyals, and Dean(2015)}]{hinton2015distilling}
Hinton, G.; Vinyals, O.; and Dean, J. 2015.
\newblock Distilling the knowledge in a neural network.
\newblock \emph{arXiv preprint arXiv:1503.02531}.

\bibitem[{Ioffe and Szegedy(2015)}]{ioffe2015batch}
Ioffe, S.; and Szegedy, C. 2015.
\newblock Batch normalization: Accelerating deep network training by reducing
  internal covariate shift.
\newblock \emph{arXiv preprint arXiv:1502.03167}.

\bibitem[{Kim et~al.(2019)Kim, Park, Na, and Yoon}]{kim2019spiking}
Kim, S.; Park, S.; Na, B.; and Yoon, S. 2019.
\newblock Spiking-yolo: Spiking neural network for real-time object detection.
\newblock \emph{arXiv preprint arXiv:1903.06530}, 1.

\bibitem[{Kim, Cho, and Hong(2020)}]{kim2020towards}
Kim, Y.; Cho, D.; and Hong, S. 2020.
\newblock Towards Privacy-Preserving Domain Adaptation.
\newblock \emph{IEEE Signal Processing Letters}, 27: 1675--1679.

\bibitem[{Kim and Panda(2020)}]{kim2020revisiting}
Kim, Y.; and Panda, P. 2020.
\newblock Revisiting batch normalization for training low-latency deep spiking
  neural networks from scratch.
\newblock \emph{Frontiers in neuroscience}, 1638.

\bibitem[{Kim and Panda(2021{\natexlab{a}})}]{kim2021optimizing}
Kim, Y.; and Panda, P. 2021{\natexlab{a}}.
\newblock Optimizing deeper spiking neural networks for dynamic vision sensing.
\newblock \emph{Neural Networks}, 144: 686--698.

\bibitem[{Kim and Panda(2021{\natexlab{b}})}]{kim2021visual}
Kim, Y.; and Panda, P. 2021{\natexlab{b}}.
\newblock Visual explanations from spiking neural networks using inter-spike
  intervals.
\newblock \emph{Scientific reports}, 11(1): 1--14.

\bibitem[{Krizhevsky and Hinton(2009)}]{krizhevsky2009learning}
Krizhevsky, A.; and Hinton, G. 2009.
\newblock Learning multiple layers of features from tiny images.
\newblock Technical Report~0, University of Toronto, Toronto, Ontario.

\bibitem[{Kundu et~al.(2020)Kundu, Venkat, Babu et~al.}]{kundu2020universal}
Kundu, J.~N.; Venkat, N.; Babu, R.~V.; et~al. 2020.
\newblock Universal source-free domain adaptation.
\newblock In \emph{Proceedings of the IEEE/CVF Conference on Computer Vision
  and Pattern Recognition}, 4544--4553.

\bibitem[{Lee et~al.(2020)Lee, Sarwar, Panda, Srinivasan, and
  Roy}]{lee2020enabling}
Lee, C.; Sarwar, S.~S.; Panda, P.; Srinivasan, G.; and Roy, K. 2020.
\newblock Enabling spike-based backpropagation for training deep neural network
  architectures.
\newblock \emph{Frontiers in Neuroscience}, 14.

\bibitem[{Lee, Delbruck, and Pfeiffer(2016)}]{lee2016training}
Lee, J.~H.; Delbruck, T.; and Pfeiffer, M. 2016.
\newblock Training deep spiking neural networks using backpropagation.
\newblock \emph{Frontiers in neuroscience}, 10: 508.

\bibitem[{Li et~al.(2021{\natexlab{a}})Li, Deng, Dong, Gong, and
  Gu}]{li2021free}
Li, Y.; Deng, S.; Dong, X.; Gong, R.; and Gu, S. 2021{\natexlab{a}}.
\newblock A free lunch from ANN: Towards efficient, accurate spiking neural
  networks calibration.
\newblock In \emph{International Conference on Machine Learning}, 6316--6325.
  PMLR.

\bibitem[{Li et~al.(2021{\natexlab{b}})Li, Guo, Zhang, Deng, Hai, and
  Gu}]{li2021differentiable}
Li, Y.; Guo, Y.; Zhang, S.; Deng, S.; Hai, Y.; and Gu, S. 2021{\natexlab{b}}.
\newblock Differentiable Spike: Rethinking Gradient-Descent for Training
  Spiking Neural Networks.
\newblock \emph{Advances in Neural Information Processing Systems}, 34.

\bibitem[{Liang, Hu, and Feng(2020)}]{liang2020we}
Liang, J.; Hu, D.; and Feng, J. 2020.
\newblock Do we really need to access the source data? source hypothesis
  transfer for unsupervised domain adaptation.
\newblock In \emph{International Conference on Machine Learning}, 6028--6039.
  PMLR.

\bibitem[{Miyato et~al.(2018)Miyato, Kataoka, Koyama, and
  Yoshida}]{miyato2018spectral}
Miyato, T.; Kataoka, T.; Koyama, M.; and Yoshida, Y. 2018.
\newblock Spectral normalization for generative adversarial networks.
\newblock \emph{arXiv preprint arXiv:1802.05957}.

\bibitem[{Miyato and Koyama(2018)}]{miyato2018cgans}
Miyato, T.; and Koyama, M. 2018.
\newblock cGANs with projection discriminator.
\newblock \emph{arXiv preprint arXiv:1802.05637}.

\bibitem[{Mopuri, Uppala, and Babu(2018)}]{mopuri2018ask}
Mopuri, K.~R.; Uppala, P.~K.; and Babu, R.~V. 2018.
\newblock Ask, acquire, and attack: Data-free uap generation using class
  impressions.
\newblock In \emph{Proceedings of the European Conference on Computer Vision
  (ECCV)}, 19--34.

\bibitem[{Nayak et~al.(2019)Nayak, Mopuri, Shaj, Radhakrishnan, and
  Chakraborty}]{nayak2019zero}
Nayak, G.~K.; Mopuri, K.~R.; Shaj, V.; Radhakrishnan, V.~B.; and Chakraborty,
  A. 2019.
\newblock Zero-shot knowledge distillation in deep networks.
\newblock In \emph{International Conference on Machine Learning}, 4743--4751.
  PMLR.

\bibitem[{Neftci et~al.(2019)}]{neftci2019surrogate}
Neftci, E.~O.; et~al. 2019.
\newblock Surrogate gradient learning in spiking neural networks.
\newblock \emph{IEEE Signal Processing Magazine}, 36: 61--63.

\bibitem[{Panda, Aketi, and Roy(2020)}]{panda2020toward}
Panda, P.; Aketi, S.~A.; and Roy, K. 2020.
\newblock Toward scalable, efficient, and accurate deep spiking neural networks
  with backward residual connections, stochastic softmax, and hybridization.
\newblock \emph{Frontiers in Neuroscience}, 14.

\bibitem[{Paszke et~al.(2017)Paszke, Gross, Chintala, Chanan, Yang, DeVito,
  Lin, Desmaison, Antiga, and Lerer}]{paszke2017automatic}
Paszke, A.; Gross, S.; Chintala, S.; Chanan, G.; Yang, E.; DeVito, Z.; Lin, Z.;
  Desmaison, A.; Antiga, L.; and Lerer, A. 2017.
\newblock Automatic differentiation in PyTorch.
\newblock In \emph{NIPS-W}.

\bibitem[{Roy, Jaiswal, and Panda(2019)}]{roy2019towards}
Roy, K.; Jaiswal, A.; and Panda, P. 2019.
\newblock Towards spike-based machine intelligence with neuromorphic computing.
\newblock \emph{Nature}, 575(7784): 607--617.

\bibitem[{Rueckauer et~al.(2017)Rueckauer, Lungu, Hu, Pfeiffer, and
  Liu}]{rueckauer2017conversion}
Rueckauer, B.; Lungu, I.-A.; Hu, Y.; Pfeiffer, M.; and Liu, S.-C. 2017.
\newblock Conversion of continuous-valued deep networks to efficient
  event-driven networks for image classification.
\newblock \emph{Frontiers in neuroscience}, 11: 682.

\bibitem[{Sengupta et~al.(2019)Sengupta, Ye, Wang, Liu, and
  Roy}]{sengupta2019going}
Sengupta, A.; Ye, Y.; Wang, R.; Liu, C.; and Roy, K. 2019.
\newblock Going deeper in spiking neural networks: Vgg and residual
  architectures.
\newblock \emph{Frontiers in neuroscience}, 13: 95.

\bibitem[{Sharmin et~al.(2020)Sharmin, Rathi, Panda, and
  Roy}]{sharmin2020inherent}
Sharmin, S.; Rathi, N.; Panda, P.; and Roy, K. 2020.
\newblock Inherent Adversarial Robustness of Deep Spiking Neural Networks:
  Effects of Discrete Input Encoding and Non-Linear Activations.
\newblock \emph{arXiv preprint arXiv:2003.10399}.

\bibitem[{Simonyan and Zisserman(2014)}]{simonyan2014very}
Simonyan, K.; and Zisserman, A. 2014.
\newblock Very deep convolutional networks for large-scale image recognition.
\newblock \emph{arXiv preprint arXiv:1409.1556}.

\bibitem[{Srivastava et~al.(2014)Srivastava, Hinton, Krizhevsky, Sutskever, and
  Salakhutdinov}]{srivastava2014dropout}
Srivastava, N.; Hinton, G.; Krizhevsky, A.; Sutskever, I.; and Salakhutdinov,
  R. 2014.
\newblock Dropout: a simple way to prevent neural networks from overfitting.
\newblock \emph{The journal of machine learning research}, 15(1): 1929--1958.

\bibitem[{Venkatesha et~al.(2021)Venkatesha, Kim, Tassiulas, and
  Panda}]{venkatesha2021federated}
Venkatesha, Y.; Kim, Y.; Tassiulas, L.; and Panda, P. 2021.
\newblock Federated learning with spiking neural networks.
\newblock \emph{IEEE Transactions on Signal Processing}, 69: 6183--6194.

\bibitem[{Wu et~al.(2018)Wu, Deng, Li, Zhu, and Shi}]{wu2018spatio}
Wu, Y.; Deng, L.; Li, G.; Zhu, J.; and Shi, L. 2018.
\newblock Spatio-temporal backpropagation for training high-performance spiking
  neural networks.
\newblock \emph{Frontiers in neuroscience}, 12: 331.

\bibitem[{Yosinski et~al.(2015)Yosinski, Clune, Nguyen, Fuchs, and
  Lipson}]{yosinski2015understanding}
Yosinski, J.; Clune, J.; Nguyen, A.; Fuchs, T.; and Lipson, H. 2015.
\newblock Understanding neural networks through deep visualization.
\newblock \emph{arXiv preprint arXiv:1506.06579}.

\bibitem[{Yuan et~al.(2019)Yuan, Tay, Li, Wang, and Feng}]{yuan2019revisit}
Yuan, L.; Tay, F.~E.; Li, G.; Wang, T.; and Feng, J. 2019.
\newblock Revisit knowledge distillation: a teacher-free framework.
\newblock \emph{arXiv preprint arXiv:1909.11723}.

\bibitem[{Zambrano et~al.(2019)Zambrano, Nusselder, Scholte, and
  Boht{\'e}}]{zambrano2019sparse}
Zambrano, D.; Nusselder, R.; Scholte, H.~S.; and Boht{\'e}, S.~M. 2019.
\newblock Sparse computation in adaptive spiking neural networks.
\newblock \emph{Frontiers in neuroscience}, 12: 987.

\bibitem[{Zheng et~al.(2020)Zheng, Wu, Deng, Hu, and Li}]{zheng2020going}
Zheng, H.; Wu, Y.; Deng, L.; Hu, Y.; and Li, G. 2020.
\newblock Going deeper with directly-trained larger spiking neural networks.
\newblock \emph{arXiv preprint arXiv:2011.05280}.

\end{thebibliography}

\end{document}